\pdfoutput=1

\documentclass[11pt]{article}

\usepackage{EMNLP2023}

\usepackage{times}
\usepackage{latexsym}

\usepackage[T1]{fontenc}

\usepackage[utf8]{inputenc}

\usepackage[noend]{algpseudocode}
\usepackage{algorithmicx,algorithm}
\usepackage{bm}
\usepackage{bbm}
\usepackage{microtype}
\usepackage{booktabs}       
\usepackage{amsmath}
\usepackage{amsfonts}
\usepackage{array}
\usepackage{bm}
\usepackage{multirow}
\usepackage{subcaption}
\usepackage{hyperref}       
\usepackage{url}            
\usepackage{graphicx} 
\usepackage{utfsym}
\usepackage{xcolor}
\usepackage{inconsolata}
\usepackage{CJKutf8}
\usepackage{pythonhighlight} 
\usepackage[normalem]{ulem}
\usepackage{inconsolata}

\title{RefGPT: Dialogue Generation \\ of GPT, by GPT, and for GPT}

\author{Dongjie Yang\textsuperscript{\rm 1}, 
    Ruifeng Yuan\textsuperscript{\rm 2},
    Yuantao Fan\textsuperscript{\rm 3},
    Yifei Yang\textsuperscript{\rm 1},\\
    \textbf{
    Zili Wang\textsuperscript{\rm 4},
    Shusen Wang\textsuperscript{\rm 4},
    Hai Zhao\textsuperscript{\rm 1, 5}} \\
    \textsuperscript{\rm 1}Shanghai Jiao Tong University, 
    \textsuperscript{\rm 2} Hong Kong Polytechnic University,\\
    \textsuperscript{\rm 3} Beijing University of Posts and Telecommunications\\
    \textsuperscript{\rm 1}\texttt{\{djyang.tony,yifeiyang\}@sjtu.edu.cn},
    \textsuperscript{\rm 2}\texttt{ruifeng.yuan@connect.polyu.hk},\\
    \textsuperscript{\rm 3}\texttt{yuantaofan@bupt.edu.cn},
    \textsuperscript{\rm 4}\texttt{\{ziliwang.do,wssatzju\}@gmail.com},\\
    \textsuperscript{\rm 5}\texttt{zhaohai@cs.sjtu.edu.cn}
    }

\begin{document}
\maketitle
\begin{abstract}
Large Language Models (LLMs) have attained the impressive capability to resolve a wide range of NLP tasks by fine-tuning with high-quality instruction data. However, collecting human-written data of high quality, especially multi-turn dialogues, is expensive and unattainable for most people. Though previous studies have used powerful LLMs to generate the dialogues automatically, they all suffer from generating untruthful dialogues because of the model hallucination. Therefore, we propose a method called RefGPT to generate enormous truthful and customized dialogues without worrying about factual errors caused by the model hallucination. RefGPT solves the model hallucination in dialogue generation by restricting the LLMs to leverage the given reference instead of reciting their own knowledge to generate dialogues. Additionally, RefGPT adds detailed controls on every utterance to enable high customization capability, which previous studies have ignored. On the basis of RefGPT, we also propose two high-quality dialogue datasets generated by GPT-4, namely \textbf{RefGPT-Fact} and \textbf{RefGPT-Code}. RefGPT-Fact is a dataset with 100k multi-turn dialogues based on factual knowledge and RefGPT-Code has 76k multi-turn dialogues covering a wide range of coding scenarios. Our code and datasets are released in \url{https://github.com/mutonix/RefGPT}.


\end{abstract}

\section{Introduction}
General chat models \citep{openai2023chatgpt, openai2023gpt4, anthropic2023claude} based on Large Language Models (LLMs) have shown the impressive capability to intention recognition and complete a variety of NLP tasks only via fine-tuning with a small amount of high-quality instruction data \citep{alpaca, vicuna2023, xu2023wizardlm}. However, such high-quality instruction datasets, especially multi-turn dialogues with instructions in vertical domains, requires enormous crowdsource workers with extensive professional knowledge to collect \citep{ouyang2022training}, where the cost is unaffordable for most people. 



Previous studies \citep{peng2023instruction, xu2023baize, UltraChat} have shown the effectiveness of prompting LLMs like GPT-3 \citep{brown2020language} to generate enormous instructions (single-turn dialogues) or multi-turn dialogues with given human-written instructions or conversation topics as seeds. However, such one-shot or few-shot methods have a common deficiency that they have the risk of generating untruthful and misleading content due to the language model hallucination \citep{openai2023gpt4, Ji_2023}. The reason why the issue of untruthfulness happens is obvious. This is because the quantity of information in seed prompts like human-written instructions or topics is not enough for being converted to the dialogue on a new topic so LLMs have to recite their own knowledge to complete such a new dialogue which may lead to the model hallucination of generating untruthful facts.

Therefore, we introduce RefGPT, a method for generating truthful and customized multi-turn dialogues utilizing the ability of powerful LLMs like GPT-3.5/GPT-4. RefGPT first provides a plain text or a document as the reference and guides the LLMs to leverage the references to generate dialogues. By providing enough information on a new topic as context, LLMs will be prompted not to rely on their own knowledge to generate the dialogues, thus resolving the hallucination issue.

After ensuring the authenticity of the dialogue, we further develop an effective prompting process for RefGPT to guide the LLMs to generate highly controllable dialogues in a specified uniform format which is easy for training. Previous studies \citep{xu2023baize, wang2022self} for automatically generating dialogues have very little control over the generated dialogues. For comparison, RefGPT enables LLMs to generate customized multi-turn dialogues with detailed controls on the structure, style, and content, which further gives diversity to the generated dialogues.

Based on the RefGPT, we also propose two new multi-turn dialogue datasets, namely \textbf{RefGPT-Fact} and \textbf{RefGPT-Code}. Both datasets have English and Chinese versions. RefGPT-Fact and RefGPT-Code consist of 100k and 76k high-quality multi-turn dialogues generated from GPT-4 separately, using the online encyclopedia websites and Github repositories as the references. As long as the content on the online encyclopedia website and Github codes is truthful and reliable, the authenticity of the generated dialogues can be maximally ensured. 

Besides the topics in RefGPT-Fact and RefGPT-Code, RefGPT has the potential to generate truthful dialogues on any topics or vertical domains if we give it relevant references. RefGPT enables such people working in a specific domain, e.g., the nuclear industry, to have a high-quality multi-turn dialogues dataset to train a chatbot specializing in such domain using their own knowledge base as the reference.

To sum up, our contributions are stated as follows:
\begin{itemize}
    \item We propose RefGPT, a method of generating truthful and customized dialogues using powerful LLMs. Given the reliable reference, RefGPT resolves LLM hallucination in dialogue generation to the greatest extent. RefGPT can also enable detailed customization in the structure, style and content of the dialogues.
    \item With RefGPT, we construct two new multi-turn dialogue datasets using GPT-4, called \textbf{RefGPT-Fact} and \textbf{RefGPT-Code}. To our best knowledge, RefGPT-Fact is one of the largest multi-turn dialogue datasets based on factual knowledge. And RefGPT-Code is the first and largest synthetic multi-turn dialogue dataset covering nearly all aspects of code scenarios. These have shown the capability of applying RefGPT to generate dialogues in any vertical domain by utilizing corresponding domain-specific documents.
    
\end{itemize}

\section{Related Work}
\subsection{LLM based Dialogue Generation}
The high-quality dialogue dataset is considered crucial for the success of current general chat models \citep{vicuna2023, köpf2023openassistant}. Due to the high cost of human annotation, previous studies have explored the effectiveness of using LLMs for dialogue generation. Self-Instruct \citep{wang2022self} presents a framework that facilitates the automatic generation of instruction data (single-turn dialogues) by leveraging existing LLMs. The procedure commences with a set of human-written seed tasks and progressively generates new instructions and responses by iteratively bootstrapping both the initial seeds and the newly produced data. Baize \cite{xu2023baize} generates multi-turn dialogues by leveraging LLMs to engage in a conversation with itself as both user and assistant based on the given seed topics. UltraChat \cite{UltraChat} follows a similar idea to Baize and adopts two separate LLM APIs in the generation, where one acts as the user and the other acts as the assistant. However, the dialogues produced by these methods are susceptible to hallucination problems and are uncontrollable. Therefore, we present RefGPT as a solution to generate dialogues with truthfulness and customization.

\subsection{Reference Based Dialogue Generation}
QA pair and dialogue generation based on references have also been widely used. One important requirement for these methods is to ensure the truthfulness of the generated QA pairs and dialogues. Previous studies \citep{ma2020zero, lewis2021paq} generate millions of high-quality QA pairs based on corpus documents using special-purpose question generation models. Dialogue inpainting \citep{dai2022dialog} extends this line of work to dialogues by transforming passages from Wikipedia into multi-turn dialogues using a masked conversational language model. In this work, we adopt a similar strategy using the LLMs that we take high-quality documents as references to ensure the truthfulness of the generated dialogues. 

\begin{figure*}[h]
	\centering
	\includegraphics[width=\textwidth]{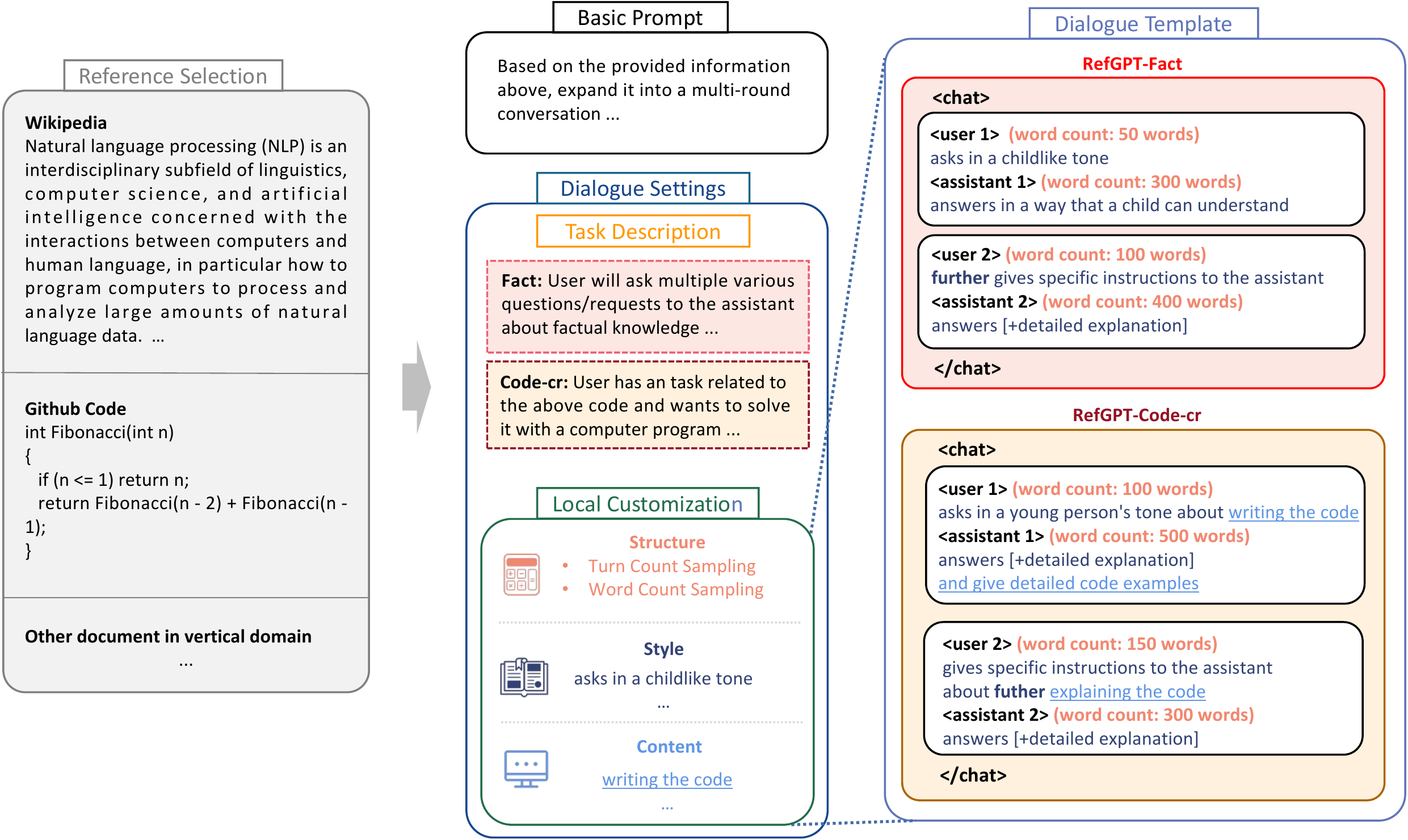}
	\caption{Overview of the whole RefGPT generation process, which mainly consists of three steps: reference selection, basic prompt and dialogue settings.}
	\label{fig:framework}
\end{figure*}

\section{Generation Process}
In this section, we present the whole process of RefGPT, which generates truthful and customized multi-turn dialogues by prompting the Large Language Models (LLMs) to effectively utilize the reference information. As illustrated in Figure \ref{fig:framework}, the RefGPT process is comprised of three main steps: Reference Selection (pertaining to truthfulness), Basic Prompt, and Dialogue Settings (pertaining to customization).



\subsection{Reference Selection}



RefGPT guides the LLMs to leverage the given external documents or plain texts as references, instead of reciting their own knowledge, to generate truthful dialogues without worrying about hallucination.

The quality of generated dialogues in RefGPT relies on the selection of appropriate references, prioritizing quality and thematic relevance. 

A reference in RefGPT can range from a piece of unrefined plain text to a high-quality and dependable document in a specific domain, whose credibility determines the upper limit of the truthfulness of the generated dialogues. On the premise that the reference has contained enough information, it is imperative to opt for high-quality references, such as authoritative knowledge-based websites like Wikipedia.

Furthermore, the chosen reference profoundly influences the thematic direction of the generated dialogues. Consequently, RefGPT exhibits the potential to generate dialogues in diverse domains, contingent upon the existence of text-based knowledge repositories within those domains. These repositories include a broad spectrum of subjects, including, but not limited to, general domains like factual knowledge with encyclopedias, program codes, and vertical domains like shopping applications or the nuclear industry.



\subsection{Basic Prompt}
To facilitate the generation of multi-turn dialogues that adhere to our basic requirements, we have devised a set of basic prompts:

\begin{enumerate}
    \item Prompt the LLMs to generate multi-turn dialogues based on the provided reference.
    \item Specify the desired language for the dialogue generation. It is preferable for the language of the reference to be consistent with the dialogue to be generated.
    \item Instruct the LLMs to reject unreasonable user requests, such as illegal or inappropriate instructions while providing appropriate advice to discourage such actions. This prompt aids in generating dialogues that align with human preferences to a certain extent.
    \item LLMs like GPT-3.5-turbo and GPT-4 offer an option of writing a "system" role prompt to exert precise control over their behaviors in responses. This capability enables customization of the chatbot's identity by providing relevant background information. For instance, in a vertical domain like a shopping app, RefGPT can generate dialogues that conform to the persona of a shopping assistant, even if the reference has no explicit association with shopping (but may have an implicit association).
\end{enumerate}

\subsection{Dialogue Settings}
\label{sec:custom}
Rather than generating dialogues uncontrollably, RefGPT uses dialogue settings to convert the reference to a specific dialogue format and customize every utterance, as shown in the middle part of the Figure \ref{fig:framework}.
In dialogue settings, we first specify the task description to tell LLMs how to use the reference. We then customize the structure, style, and content of the dialogue, which can be collectively called local customization.


\subsubsection{Task Description}

We begin by defining the task of dialogue generation concerning the utilization of references, as it relies on the specific aspect of the reference that we aim to initiate the dialogue. For instance, a given piece of program code can lead to multiple scenarios (tasks), such as explaining, creating, or debugging.


\subsubsection{Local Customization}
As per the task description, the local customization specifies the settings regarding the dialogue's structure, style, and content. These settings are then incorporated into a dialogue template for generating the final dialogue.

\paragraph{Dialogue Structure}
To define the dialogue structure, we start the dialogue with the marker \texttt{<chat>} and end it with the marker \texttt{</chat>}. These two markers specify the range of the whole dialogue. Between the start and the end, we use \texttt{<user>} for the user giving instructions and \texttt{<assistant>} for the chatbot. A unified output format in a dialogue template avoids most of the weird generations of LLMs and is easier for post-processing. What is more, we will show more merits of using such a format to control the number of turns and length per turn.

\quad
\paragraph{(1) Number of Turns}
LLMs like GPT-3.5/GPT-4 often fail with counting the number of the turns of dialogues if we directly require a certain number. But we find that GPT-3.5/GPT-4 are good at following the given format and replacing the placeholders with their own generated content. Therefore, if we want to generate $n$ turns of dialogues, we explicitly give the $n$ \texttt{<user>} and \texttt{<assistant>} pairs to let LLMs follow the output format. We have also added numerical markers to indicate the $i^{th}$ turn of the dialogue, e.g., \texttt{<user $i$>} and \texttt{<assistant $i$>}, allowing the LLMs to better identify the progress of the current generated turn.

\quad
\paragraph{(2) Length of Utterance}
Generating a whole dialogue at one-time, e.g., Self-Instruct \citep{wang2022self}, often leads to much shorter responses than the general chat models like GPT-3.5 do, as shown in Table \ref{tab:dataset}. However, in RefGPT, we can control the lengths of not only the responses of the assistant but also the questions raised by the user at every turn of the dialogue.

 We observe that specifying a word count as the prompt is useful for influencing the length of generated utterances. Following the auto-regressive (left-to-right) order, we first illustrate the requirement of word count like \texttt{<user>(word count: x words)} or  \texttt{<assistant>(word count: x words)} before our customization on style and content. Therefore, RefGPT can generate a shorter or much longer question/response depending on the specified word count. Though this prompt can also be used to make the generated utterances longer with other methods like Self-Instruct, generating longer utterances always leads to a more severe hallucination problem. RefGPT filters out the reference whose length is shorter than 80\% of the required dialogue length to ensure truthfulness. Thus the LLMs have no necessity of reciting their own knowledge, as the reference length is similar and even longer than the dialogue length. 


\paragraph{Dialogue Style} Staying organized around the same reference, the style of dialogue can vary in the style of asking and answering. For example, a dialogue can start between a user who is a child and an assistant who answers in a way that a child can understand. RefGPT enables this customization for every utterance of \texttt{<user>} and \texttt{<assistant>} in the dialogues by adding the style requirements before the content customization.

\paragraph{Dialogue Content}

After specifying the style, we can customize the content of each utterance about what to ask and what to answer.

For the task like factual knowledge, the user can be set to ask more about the entity or numbers in the reference. For the task of coding, the user can ask from different perspectives on writing, revising, and using the code and the assistant can choose to give an example or not.

\paragraph{Dialogue Template}
We aggregate the local customizations into a dialogue template to transfer the reference to the dialogue. To enable diversity, we sample different local customization settings for each utterance in the dialogue, as shown in the right-most part in Figure \ref{fig:framework}. In practice, RefGPT can work well even without style and content pools. These additional settings only need a small amount of manual work for further customization and can be reused to generate diverse dialogues based on different references.

\begin{enumerate}
    \item For the dialogue structure, we will set the number of turns by weighted sampling. And we sample the word count for both user and assistant in each utterance from a Gaussian distribution.
    \item For the dialogue style, we construct a conversational style pool to sample the style settings.
    \item For the dialogue content, we construct a content pool according to the task (factual knowledge, code, etc) to sample the content settings.
\end{enumerate}


\begin{table*}[ht]
 \setlength{\tabcolsep}{4.3pt}
 \centering
\caption{Comparsions on different dialogue datasets that contain instructions. \textbf{AI} means whether it is generated by AI. \textbf{Truthful} indicates whether the truthfulness of the dialogues is guaranteed. \textbf{QLen} means the average number of tokens\footnotemark of user utterance.  \textbf{RLen} means the average number of tokens of assistant utterance. \textbf{Turn} means whether the number of dialogue turns can be specified. \textbf{Lang} indicates the languages the dataset supports. For a fair comparison, only the English parts are selected in all the datasets.}
\label{tab:dataset}
\vskip 0.1in \small
\begin{tabular}{lcccccc}
\toprule
\textbf{Dataset}  & AI & \textbf{Truthful} & \textbf{QLen} & \textbf{RLen} & \textbf{Turn} & \textbf{Lang}  \\
\midrule
  Dolly \cite{data2023dolly} & \color{red}\usym{2718} & N/A & 16.3  &  78.2 & 1 & en \\
  Oasst1 \citep{köpf2023openassistant} & \color{red}\usym{2718} & N/A & 28.0 & 169.5 & 1$\sim$5&  multi \\
  ShareGPT \citep{dom2023sharegpt} & \color{green}\usym{2714} & \color{red}\usym{2718} & 75.6 & 268.8 & 1$\sim$5 & multi \\
  Alpaca \citep{wang2022self}  &  \color{green}\usym{2714} & \color{red}\usym{2718} & 17.2 & 55.3  & 1  & en \\
  Baize Quora \citep{xu2023baize}  & \color{green}\usym{2714} & \color{red}\usym{2718} & 15.7 & 43.2 & 3$\sim$5 & en \\
  UltraChat World \citep{UltraChat}  &\color{green}\usym{2714} & \color{red}\usym{2718} &  28.6  &  207.9 & 3$\sim$7 &  en\\
  RefGPT-Fact & \color{green}\usym{2714} & \color{green}\usym{2714} & 28.1 & 269.5 & 3$\sim$4 & en, cn \\
  RefGPT-Code-ds & \color{green}\usym{2714} & \color{green}\usym{2714} & 281.7 & 374.6 & 3$\sim$4 & en, cn \\
  RefGPT-Code-cr & \color{green}\usym{2714} & \color{green}\usym{2714} & 36.9 & 395.0 & 3$\sim$4 & en, cn\\
  RefGPT-Code-bg & \color{green}\usym{2714} & \color{green}\usym{2714} & 155.7 & 380.8 & 2$\sim$4 & en, cn\\
\bottomrule
\end{tabular}
\end{table*}
\footnotetext{In Table \ref{tab:dataset}, we calculate the number of tokens with the OpenAI tokenizer library of \texttt{tiktoken} in \url{https://github.com/openai/tiktoken}. We use \texttt{cl100k\_base} in \texttt{tiktoken} to tokenize.}

\section{RefGPT Dialogue Datasets}
In this section, we present two multi-turn dialogue datasets, denoted as \textbf{RefGPT-Fact} and \textbf{RefGPT-Code}, which are generated utilizing the GPT-4 API in conjunction with RefGPT. More information about these two datasets can be found in Appendix \ref{app:dataset}, and examples are provided in Appendix \ref{app:example}.

\subsection{Dataset Generation Process}
\paragraph{RefGPT-Fact} RefGPT-Fact is a dataset containing 100k multi-turn dialogues about factual knowledge with 50k English and 50k Chinese. The English version uses the English Wikipedia as the reference and the Chinese version uses the frequently-used Chinese online encyclopedia website, Baidu Baike. We use various dialogue settings mentioned in Sec \ref{sec:custom} to increase the dialogue diversity.

\paragraph{RefGPT-Code} RefGPT-Code is a dataset containing 76k multi-turn dialogues about programming with 37k English and 39k Chinese, which has covered most aspects of code usage scenarios and multiple types of programming languages. Both the English version and Chinese version use the public Github dataset on Google BiqQuery with no overlap in these two languages. RefGPT-Code has derived various ways of leveraging the program code as the reference to enable different scenarios. We consider three perspectives of code discussion, code creation and bug fixing in RefGPT-Code.

\begin{enumerate}
    \item In \textbf{RefGPT-Code-ds} about code discussion, we want the LLMs to generate dialogues about asking questions about the given reference code, including explaining, discussing, revising, rewriting, and using the code. After the generation, we will concatenate the reference code as the context to the first question of the user to form the complete version of the dialogue, because we often give the code first before asking questions about it. Thus, the whole dialogue has much longer user utterances, as shown in Table \ref{tab:dataset}.
    \item In \textbf{RefGPT-Code-cr} about code creation, though we provide the program code as the reference, we assume that the user has an idea/request/trouble/task relevant to the given code but does not know such a code exists, thus he/she wants the assistant to help with writing the code. And the assistant is required to write the code according to the reference code instead of generating a new code to ensure the reliability of the generated code.
    \item  In \textbf{RefGPT-Code-bg} about bug fixing, the user first writes a piece of code with bugs based on the given reference code, which is realized by asking the LLMs to rewrite the code to a buggy one in the first utterance of the user. Then the assistant is required to tell the user where the bugs are and how to fix them according to the reference code. In this scenario, we assume the reference code is reliable and has no bugs.
\end{enumerate}

\subsection{Dataset Collection Setup}
We use the RefGPT with GPT-4 API to generate these two datasets.
The length of every utterance is decided by sampling the Gaussian distribution of $\mathcal{N}(\mu, \sigma)$, where $\mu$ accounts for the average word count (e.g., 300 words) of the utterance and $\sigma$ is the standard variance (e.g., 50 words). The number of turns is decided by weighted sampling, where the weights determine the ratio of dialogues with a specific number of turns in the dataset.


\subsection{Dataset Statistics}
As shown in Table \ref{tab:dataset}, we compare our datasets to other high-quality dialogue datasets. ShareGPT \citep{dom2023sharegpt} collects the dialogues from the real users and ChatGPT, which have much longer user utterances and assistant utterances. If we choose the responses of ChatGPT as a baseline, methods with one API, e.g., Self-Instruct \citep{wang2022self} and Baize \citep{xu2023baize}, always lead to shorter assistant responses. UltraChat \citep{UltraChat} with two independent APIs chatting to each other maintains the length of generated responses close to ChatGPT. However, as shown in Table \ref{tab:method}, such methods call the model API one utterance at a time with significantly increasing cost and time, as UltraChat has to attach the conversation history multiple times. By contrast, RefGPT generates the whole dialogue with one API call but can adjust the length of generated utterance flexibly according to the requirement. 

RefGPT-Fact inherits the diversity of the references like Wikipedia and Baidu Baike. Besides that, RefGPT-Fact has an average response length of 269.5 in English which is very similar to the length of ChatGPT response in ShareGPT. 

RefGPT-Code series implements various customizations to be adapted to specific scenarios and have longer user and assistant utterances because we have not only the utterances but also the code attached to the dialogues.


\begin{table*}[ht]
 \setlength{\tabcolsep}{4.3pt}
 \centering
\caption{Comparsions on different methods of automatically generating dialogues via LLMs. \textbf{Multi-turn} means whether it is a multi-turn dialogue generation. \textbf{Human Tru.}  and \textbf{GPT-4 Tru.} evaluate the truthfulness with accuracy by humans and GPT-4 model. \textbf{Len} uses ChatGPT's response length as the standard for long responses. \textbf{Turn} means whether the number of dialogue turns can be specified. \textbf{Custo.} depends on whether it can control the dialogue structure and content. \textbf{Call} is the number of model or model API calls needed for generating an instruction or a $n$-turn dialogue every time.}
\label{tab:method}
\vskip 0.1in \small
\begin{tabular}{lccccccc}
\toprule
\textbf{Method} & \textbf{Multi-turn} & \textbf{Human Tru.} &\textbf{GPT-4 Tru.} & \textbf{Len} & \textbf{Turn} & \textbf{Custo.} & \textbf{Call} \\
\midrule
  Self-Instruct \citep{wang2022self} & \color{red}\usym{2718} & 54.0 & 50.2 & short & one & limited & 1 \\
  Baize Self-Chat \citep{xu2023baize} & \color{green}\usym{2714} & 50.0 &  47.2 & short & random &limited  & 1 \\
  UltraChat \citep{UltraChat} & \color{green}\usym{2714} & - & - & long & adjustable & limited & $2n$\\
  RefGPT & \color{green}\usym{2714} & 98.0 & 97.5 & adjustable &adjustable & highly &  1 \\
\bottomrule
\end{tabular}
\end{table*}

\section{Experiment}
\subsection{Truthfulness Evaluation}
\label{sec:trutheval}
In order to verify the reliability of RefGPT, We evaluate the truthfulness of the RefGPT dataset using both human evaluation for small sample and automatic evaluation with GPT-4 for a large range of verificaiton. For automatic evaluation with GPT-4, though existing methods \citep{vicuna2023, liu2023geval} have leveraged the GPT-4 to evaluate the performance of other LLMs. However such evaluation is not reliable in factual error checking because GPT-4 has the issue of model hallucination. Inspired by RefGPT, we design a pipeline to evaluate the truthfulness of generated dialogues from our reference datasets, e.g., Wikipedia, by using the GPT-4 to evaluate but with the additional help of reference.


\begin{figure}[h]
	\centering
	\includegraphics[width=0.47\textwidth]{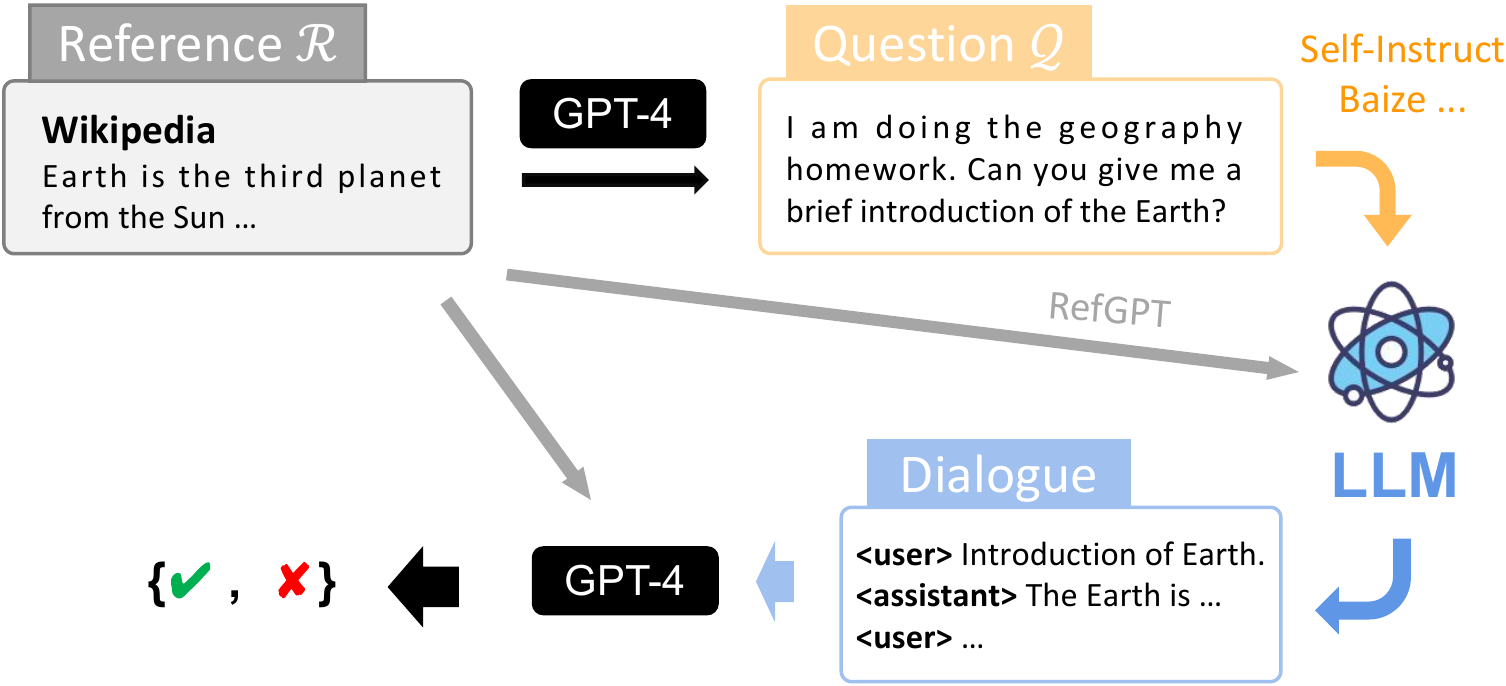}
	\caption{Illustration of the process of truthfulness evaluation.}
	\label{fig:truthfuleval}
\end{figure}

\begin{table*}[ht]
 \centering
\caption{Examples of truthfulness evaluation on the generated dialogues by RefGPT. We prompt GPT-4 to give the explanation first and then output the judgment.}
\label{tab:eval_example}
\begin{tabular}{p{4.1cm}p{5.3cm}p{3.8cm}cp{0.8cm}}
\toprule

\textbf{Reference} & \textbf{Dialogue} & \textbf{Explanation} & \textbf{Result} \\
\midrule
\small{R. J. Hampton Jr ... announced that he was transferring to the nationally-ranked basketball program of Montverde Academy in Montverde, Florida, before deciding to return to Little Elm ...} &

\small{\textbf{<user>}  I heard that R. J. Hampton Jr. is one of the talented basketball players in the NBA. Can you tell me more about his career and background? }

\small{\textbf{<assistant>} ... he transferred to Montverde Academy, where he continued to excel and earn multiple honors ...}&
\small{R. J. Hampton Jr did not transfer to Montverde Academy. He had announced his intention to transfer but ultimately decided to return to Little Elm High School ... } & \color{red}\usym{2718} \\
\midrule
\small{Star Wars: Rebel Assault is a rail shooter video game developed and published by LucasArts ...}&
\small{\textbf{<user>} What is Star Wars: Rebel Assault?}

\small{\textbf{<assistant>} Star Wars: Rebel Assault is a famous rail shooter video game, which was developed and published by LucasArts ...} &
\small{Assistant's answers accurately reflect the information provided, covering the game's significance, storyline, gameplay mode ...} & \color{green}\usym{2714} \\
\bottomrule
\end{tabular}
\end{table*}

\subsubsection{Evaluation Process}
We compare RefGPT to two popular automatic methods as the baselines, namely Self-Instruct \citep{wang2022self} and Baize Self-Chat \citep{xu2023baize}. For a fair comparison, we want the generated dialogues of different methods to talk about the same things. Thus we do an additional work that we let GPT-4 generate \{question, answer\} pairs from the selected references and restrict the answers to the questions to be found or inferred from the references. Given a selected reference, for Self-Instruct, we follow the Alpaca \cite{alpaca} that we randomly select three \{question, answer\} pairs (from other references) as few-shot examples and add the final question from the selected reference at the end of the model input. And we let the model respond to the final question. For Baize, we use the question generated from the selected reference as the seed following the way that Baize uses questions in Quora as seeds. For RefGPT, we directly use the selected reference to generate. In practice, we select 1000 passages from Wikipedia as the references to generate 1000 seed \{question, answer\} pairs using the GPT-4. And we generate the dialogues using these three methods with GPT-3.5-turbo for the experiment. In Table \ref{tab:eval_example}, we give examples of evaluating the truthfulness of the dialogues generated by RefGPT. And more examples of Self-Instruct and Baize can be seen in Appendix \ref{app:truthfulness}.

For human evaluation for a small sample, we randomly sample 50 English dialogues each for Alpaca, Baize, and RefGPT about factual knowledge. And 2 humans evaluate the truthfulness of the dialogues according to the references. 

For automatic evaluation for a large range, in order to let GPT-4 check the factual errors without suffering from the model hallucination, we need a reference for GPT-4 to refer like RefGPT. Therefore, as shown in Figure \ref{fig:truthfuleval}, we let GPT-4 check if the generated dialogue accords with the reference. If the generated dialogue does not align with the reference, it indicates the presence of factual errors.


\subsubsection{Result}
We use accuracy to measure the truthfulness in the evaluation process, which is the proportion of the number of dialogues without factual errors in the total of 1000 generated dialogues. In Table \ref{tab:method}, to our surprise, we can see that Self-Instruct and Baize Self-Chat have a striking number of factual errors in the generated dialogues on both human and GPT-4 evaluations. As the dialogues generated by Baize are multi-turn, they are more likely to contain factual errors and thus have a lower truthfulness score of 47.2. By contrast, RefGPT has a truthfulness score of 97.5 with merely no factual errors. This also implicitly indicates that a model like GPT-3.5-turbo already has the ability to generate the dialogues strictly conforming to the references rather than modifying with hallucination. Another method called UltraChat \citep{UltraChat} in Table \ref{tab:method} is not included, as the code has not been open-source at the time we write this paper. 

\subsection{Further Analysis}
 In this section, we explore the potential influence of the reference and customization on the generated dialogues by RefGPT. For each setting in the following experiments, we generate 1000 dialogues using GPT-3.5-turbo.
\subsubsection{Dialogue Quality}
As RefGPT generates the dialogue according to the reference, the reference has a significant impact on the quality of generated dialogues. We use the evaluation method mentioned in Sec \ref{sec:trutheval} to evaluate the influence of the dialogue quality (truthfulness) in the following validations.

\begin{table}[h]
 \setlength{\tabcolsep}{4.3pt}
 \centering
\caption{The truthfulness scores of 3-turn dialogues with different ratios of reference length and noise levels.}
\label{tab:analysis_quality}
\vskip 0.1in \small
\begin{tabular}{lc|lc}
\toprule

\textbf{Ref. Ratio} & \textbf{Truthfulness} &\textbf{Noise Level} & \textbf{Truthfulness} \\
\midrule
  100\% & 96.5 & 0\% & 96.5 \\
  50\%  & 96.2 & 10\% & 96.2\\
  25\%  & 97.3 & 20\% & 94.8 \\
\bottomrule
\end{tabular}
\end{table}

\paragraph{Reference Length} As length is proportional to the amount of information the reference contains, we want to find out how the reference length will influence the truthfulness of the generated dialogues. We use the dialogue template of a 3-turn dialogue, where each utterance word count of the assistant is required to be 300 words. We experiment on different lengths of reference by the proportions: 100\%, 50\%, and 25\% of the original required length (3 $\times$ 300 = 900 words).

As shown in Table \ref{tab:analysis_quality}, it is surprising to see that the truthfulness scores do not decrease much as the reference lengths are greatly reduced. We find that the GPT-3.5-turbo chooses to reduce the length of the generated utterances to obey reference despite violating the length requirement.

\paragraph{Reference Quality} The reference in RefGPT can vary from plain texts to cleaned documents in the vertical domain. 

In order to quantify the influence of reference quality on dialogue quality, we experiment with different qualities of references by adding additional noise. To be specific, we use the original reference as the baseline. We use HTML labels as noise is that many references may come from the crawled data on the websites and contain many HTML labels as noise if we do not clean the data carefully. We experiment with adding 10\% and 20\% nonsense HTML labels as the noise.

As we can see in Table \ref{tab:analysis_quality}, the truthfulness of the generated dialogues only slightly decreases because of the additional noise. This indicates the good robustness of generating truthful dialogues even with GPT-3.5-turbo.

\subsubsection{Dialogue Structure}
\label{sec:eval_struct}
During post-processing of the generated dialogues of RefGPT, we find that the input length (related to reference length) and output length (related to the required word count) will influence the success rate of obeying the dialogue template. In order to evaluate the customization ability of RefGPT, we do experiments on generating 3-turn and 5-turn dialogues. As the input length (reference length) is also determined by the required word count, we experiment with different word counts of 100, 300, and 600 for each assistant utterance to verify the success rate of obeying the dialogue template. 


\begin{table}[h]
 \setlength{\tabcolsep}{4.3pt}
 \centering
\caption{The success rates (\%) of obeying the dialogue templates with different word count settings for 3-turn and 5-turn dialogues.}
\label{tab:ablation_structure}
\vskip 0.1in \small
\begin{tabular}{lccc}
\toprule

\textbf{Word Count} & \textbf{Turn} &\textbf{w/ \texttt{</chat>}} & \textbf{w/o \texttt{</chat>}} \\
\midrule
  100 & 3 / 5  & 97.4 / 94.8 & 93.5 / 91.4 \\
  300 & 3 / 5 & 94.6 / 90.1  & 91.3 / 88.5\\
  600 & 3 / 5 & 93.2 / 86.5 & 88.4 / 70.4 \\
\bottomrule
\end{tabular}
\end{table}

From Table \ref{tab:ablation_structure}, we can see that dialogues with fewer tokens to generate (fewer words in assistant utterances and fewer turns) will lead to better control over the dialogue structure with a higher success rate. We further observe that if the ending mark \texttt{</chat>} is successfully generated, the dialogues are more likely to obey the dialogue template with the correct number of turns.

\section{Conclusion}
We present RefGPT, a new method that generates truthful and customized multi-turn dialogues using LLMs like GPT-3.5/GPT-4. Incorporating a reliable reference, RefGPT minimizes hallucination and untruthful content generation. RefGPT also allows for dialogue customization in structure, style, and content, making it flexible to generate dialogues with diversity. On the basis of RefGPT, we also use GPT-4 to construct two new multi-turn dialogue datasets, RefGPT-Fact and RefGPT-Code, based on the online encyclopedia websites and Github repositories. These datasets also showcase RefGPT's significant potential for developing dependable, domain-specific dialogue data required by specialized chatbots and other natural language processing applications.

\section*{Limitations}
RefGPT can only strictly generate the dialogues conforming to the references even though the reference itself may have factual errors. Furthermore, the generated dialogues can not be avoided to be influenced by the biases from the references. Thus the datasets RefGPT-Fact and RefGPT-Code may have factual errors and typos from Wikipedia, or bugs and malicious program codes from Github repositories. 

LLMs like GPT-3.5/GPT-4 have their own biases, which will also have reflections in the dialogues generated by RefGPT.


\bibliography{custom}
\bibliographystyle{acl_natbib}
\clearpage
\appendix
\onecolumn






\section{Dataset Card}
\label{app:dataset}
\subsection{RefGPT-Fact}
\label{app:fact}
RefGPT-Fact is a dataset comprising 100k multi-turn dialogues focusing on factual knowledge. There are two versions, with the English version containing 50k dialogues based on the English Wikipedia, while the Chinese version consists of 50k dialogues sourced from the widely-used Chinese online encyclopedia, Baidu Baike.

Since most of the passages in the English Wikipedia and Baidu Baike are written by individuals or unofficial organizations, many of the passages are not commonly seen in everyday life. We use GPT-3.5-turbo API to quickly filter out the uncommon passages by asking it "\texttt{Do you know xxx? If yes, return <yes>. If no, return <no>.}", where \texttt{xxx} is the title of the passage\footnote{This method is based on knowledge of GPT-3.5-turbo, where recall rate is limited. A more recommended way is using the access rate to filter out the uncommon passages.}.

\subsection{RefGPT-Code}
\begin{figure}[h]
	\centering
	\includegraphics[width=0.45\textwidth]{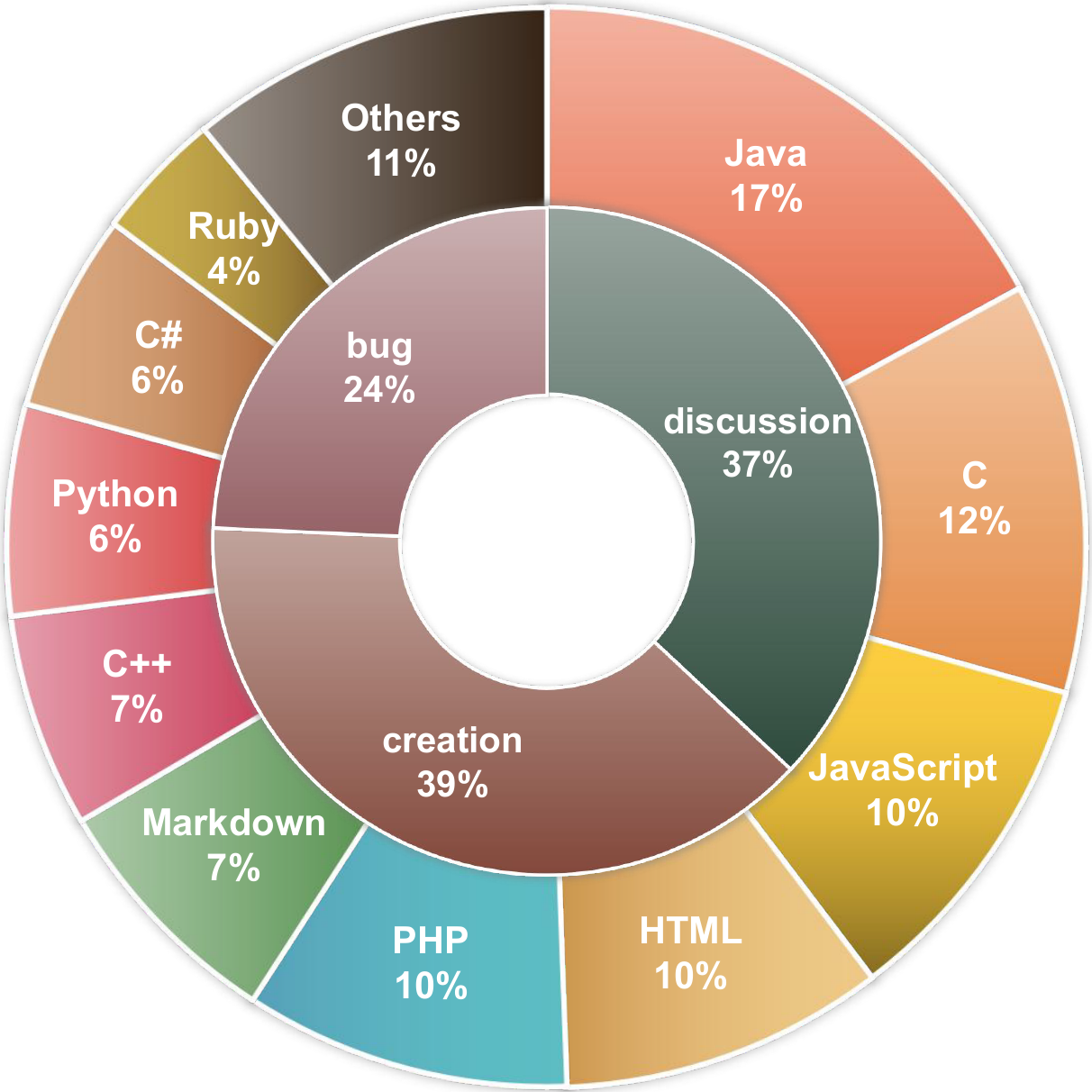}
	\caption{Composition of RefGPT-Code Dataset including English and Chinese.}
	\label{fig:code}
\end{figure}

RefGPT-Code is a comprehensive dataset that consists of 76k multi-turn dialogues on programming, including 37k English and 39k Chinese dialogues. As illustrated in Figure \ref{fig:code}, it encompasses a wide range of coding scenarios about discussion, creation, and bug fixing using various programming languages. The dataset utilizes the public Github dataset available on Google BigQuery, with no overlapping data between the two languages. 
\label{app:code}

\begin{table*}[h]
 \centering
\caption{An example of the prompt for generating the English RefGPT-Fact data. \textbf{\{dialogue\_template\}} is replaced by the dialogue template mentioned in Sec \ref{sec:custom}.}
\begin{tabular}{p{15cm}}
\toprule
\texttt{\small{\#\#Provided Information\#\# \textbf{\{reference\}}\quad Based on the \#\#Provided Information\#\# above and its relevant topic, expand it into a multi-round conversation. The conversation requires you to act as the chatbot Assistant and interact with a human, helping to solve the requests raised by the human. The human will ask multiple various questions/requests to the Assistant based on the information above (but the conversation should not include expressions like "according to the above information"), and the subsequent questions/requests will be a follow-up based on the previous conversation history. For every reasonable question/request posed by Human, Assistant should provide as detailed an answer as possible, offering further explanations or examples. For unreasonable requests from Human (those that are harmful to society, immoral, or illegal), Assistant will refuse to answer and explain the reason for not answering, while also providing reasonable advice to avoid such actions.}}

\texttt{\small{\#Conversation Plan\# Example: "<chat><Human 1>:(Word count requirement: x words)XXX <Assistant 1>: (Word count requirement: x words) XXX <Human 2>:(Word count requirement: x words)XXX <Assistant 2>: (Word count requirement: x words) XXX </chat>", "XXX" is the requirement for the current conversation content of that role, and "(Word count requirement: x words)" specifies the minimum word count requirement for utterance of Human or Assistant. It must be noted: the conversation starts with <chat> as the beginning of the multi-round conversation and ends with </chat> as the end of the multi-round conversation.
The following conversation follows this \#Conversation Plan\# and word count requirements: "\textbf{\{dialogue\_template\}}", a total of \textbf{\{number\_of\_turns\}} turns of conversation.}}
\\
\midrule
\texttt{\small{\textbf{\{dialogue\_template\}}\quad<chat><Human 1>:(word count: 100 words)asks a question <Assistant 1>:(word count: 200 words)answers [+detailed explanation] <Human 2>:(word count: 150 words)further asks from the perspective of real life <Assistant 2>:(word count: 100 words)answers [+detailed explanation] <Human 3>:(word count: 50 words)further asks a question <Assistant 3>:(word count: 150 words)answers [+detailed explanation] </chat>}}\\
\bottomrule
\end{tabular}
\end{table*}

\begin{table*}[h]
 \centering
\caption{An example of the prompt for generating the English RefGPT-Code-ds data.}
\begin{tabular}{p{15cm}}
\toprule
\texttt{\small{\#\#Provided Information\#\# \textbf{\{reference\}}\quad Based on the \#\#Provided Information\#\# above and its relevant topic, expand it into a multi-round conversation. The conversation requires you to act as the chatbot Assistant and interact with a human, helping to solve the requests raised by the human. The human will ask multiple various questions/requests to the Assistant based on the information above (but the conversation should not include expressions like "according to the above information"), and the subsequent questions/requests will be a follow-up based on the previous conversation history. For every reasonable question/request posed by Human, Assistant should provide as detailed an answer as possible, offering further explanations or examples. For unreasonable requests from Human (those that are harmful to society, immoral, or illegal), Assistant will refuse to answer and explain the reason for not answering, while also providing reasonable advice to avoid such actions.}}

\texttt{\small{\#Conversation Plan\# Example: "<chat><Human 1>:(Word count requirement: x words)XXX <Assistant 1>: (Word count requirement: x words) XXX <Human 2>:(Word count requirement: x words)XXX <Assistant 2>: (Word count requirement: x words) XXX </chat>", "XXX" is the requirement for the current conversation content of that role, and "(Word count requirement: x words)" specifies the minimum word count requirement for utterance of Human or Assistant. It must be noted: the conversation starts with <chat> as the beginning of the multi-round conversation and ends with </chat> as the end of the multi-round conversation.
The following conversation follows this \#Conversation Plan\# and word count requirements: "\textbf{\{dialogue\_template\}}", a total of \textbf{\{number\_of\_turns\}} turns of conversation.}}
\\
\midrule
\texttt{\small{\textbf{\{dialogue\_template\}}\quad<chat><Human 1>:(word count: 50 words)makes a request \uline{about writing the code} <Assistant 1>:(word count: 250 words)answers [+detailed explanation] \uline{and give code examples} <Human 2>:(word count: 100 words)asks in a young person's tone \uline{about further modifying the code} <Assistant 2>:(word count: 300 words)answers [+detailed explanation] \uline{and give code examples} <Human 3>:(word count: 20 words)asks from the perspective of real life \uline{about further how to use the code} <Assistant 3>:(word count: 250 words)answers [+detailed explanation] \uline{and give code examples} </chat>}}\\

\bottomrule
\end{tabular}
\end{table*}

\begin{table*}[h]
 \centering
\caption{An example of the prompt for generating the English RefGPT-Code-cr data.}
\begin{tabular}{p{15cm}}
\toprule
\texttt{\small{\#\#Provided Information\#\# \textbf{\{reference\}}\quad Based on the \#\#Provided Information\#\# above and its relevant topic, expand it into a multi-round conversation. \uline{Human has an idea / requirement / task / assignment / problem / difficulty related to the above code and wants to solve it with a computer program, but doesn't know how to do it. But Human doesn't know that the above code exists, so it can't be mentioned in conversation. Assistant needs to organize the above code into answers (which cannot be found by Human) according to Human's ideas, write specific program code for Human and explain it in detail so that Human's ideas can be realized. Based on this idea, Human would ask multiple questions and requests for specific code written by the Assistant, which will be follow-ups based on the previous conversation history.} For unreasonable requests from Human (those that are harmful to society, immoral, or illegal), Assistant will refuse to answer and explain the reason for not answering, while also providing reasonable advice to avoid such actions.}}

\texttt{\small{\#Conversation Plan\# Example: "<chat><Human 1>:(Word count requirement: x words)XXX <Assistant 1>: (Word count requirement: x words) XXX <Human 2>:(Word count requirement: x words)XXX <Assistant 2>: (Word count requirement: x words) XXX </chat>", "XXX" is the requirement for the current conversation content of that role, and "(Word count requirement: x words)" specifies the minimum word count requirement for utterance of Human or Assistant. It must be noted: the conversation starts with <chat> as the beginning of the multi-round conversation and ends with </chat> as the end of the multi-round conversation.
The following conversation follows this \#Conversation Plan\# and word count requirements: "\textbf{\{dialogue\_template\}}", a total of \textbf{\{number\_of\_turns\}} turns of conversation.}}
\\
\midrule
\texttt{\small{\textbf{\{dialogue\_template\}}\quad<chat><Human 1>:(word count: 50 words)asks with curiosity \uline{about creating the code} <Assistant 1>:(word count: 300 words)answers [+detailed explanation] \uline{and give code examples} <Human 2>:(word count: 100 words)asks a question \uline{about further using the code} <Assistant 2>:(word count: 250 words)answers [+detailed explanation] \uline{and give code examples} <Human 3>:(word count: 150 words)asks a question \uline{about further explaining the code} <Assistant 3>:(word count: 300 words)answers [+detailed explanation] \uline{and give code examples} <Human 4>:(word count: 50 words)expresses his/her needs and asks the Assistant for help \uline{about further using the code} <Assistant 4>:(word count: 200 words)answers [+detailed explanation]</chat>}}\\
\bottomrule
\end{tabular}
\end{table*}

\begin{table*}[h]
 \centering
\caption{An example of the prompt for generating the English RefGPT-Code-bg data.}
\begin{tabular}{p{15cm}}
\toprule
\texttt{\small{\#\#Provided Information\#\# \textbf{\{reference\}}\quad Based on the \#\#Provided Information\#\# above and its relevant topic, expand it into a multi-round conversation. \uline{ Human will write a piece of code with bugs based on the given code above (however, Human needs to hide the presence of the given code in the conversation, and it cannot be mentioned). They will then ask Assistant for help in fixing the bugs. Assistant needs to identify the mistakes in Human's code based on the given code above (but given code cannot be discovered by Human, and it cannot be mentioned in the conversation) and provide detailed explanations on how to fix the bugs, along with more explanations or examples if necessary. Afterward, Human and Assistant will continue the conversation around this code.} For unreasonable requests from Human (those that are harmful to society, immoral, or illegal), Assistant will refuse to answer and explain the reason for not answering, while also providing reasonable advice to avoid such actions.}}

\texttt{\small{\#Conversation Plan\# Example: "<chat><Human 1>:(Word count requirement: x words)XXX <Assistant 1>: (Word count requirement: x words) XXX <Human 2>:(Word count requirement: x words)XXX <Assistant 2>: (Word count requirement: x words) XXX </chat>", "XXX" is the requirement for the current conversation content of that role, and "(Word count requirement: x words)" specifies the minimum word count requirement for utterance of Human or Assistant. It must be noted: the conversation starts with <chat> as the beginning of the multi-round conversation and ends with </chat> as the end of the multi-round conversation.
The following conversation follows this \#Conversation Plan\# and word count requirements: "\textbf{\{dialogue\_template\}}", a total of \textbf{\{number\_of\_turns\}} turns of conversation.}}
\\
\midrule
\texttt{\small{\textbf{\{dialogue\_template\}}\quad<chat><Human 1>:(word count: 500 words)asks from the perspective of real life \uline{about writing a piece of code with bugs and show the detailed code} <Assistant 1>:(Word count: 250 words)answers [+detailed explanation] \uline{and tell Human about the error location in the code, then provide a correct piece of code} <Human 2>:(word count: 100 words)makes a request \uline{about further using the code} <Assistant 2>:(Word count: 200 words)answers [+detailed explanation] \uline{and give code examples} <Human 3>:(word count: 50 words)asks with curiosity \uline{about further explaining the code} <Assistant 3>:(Word count: 250 words)answers [+detailed explanation]</chat>}}\\
\bottomrule
\end{tabular}
\end{table*}

\clearpage
\section{Dataset Examples}
\label{app:example}
\begin{figure*}[h]
	\centering
	\includegraphics[width=\textwidth]{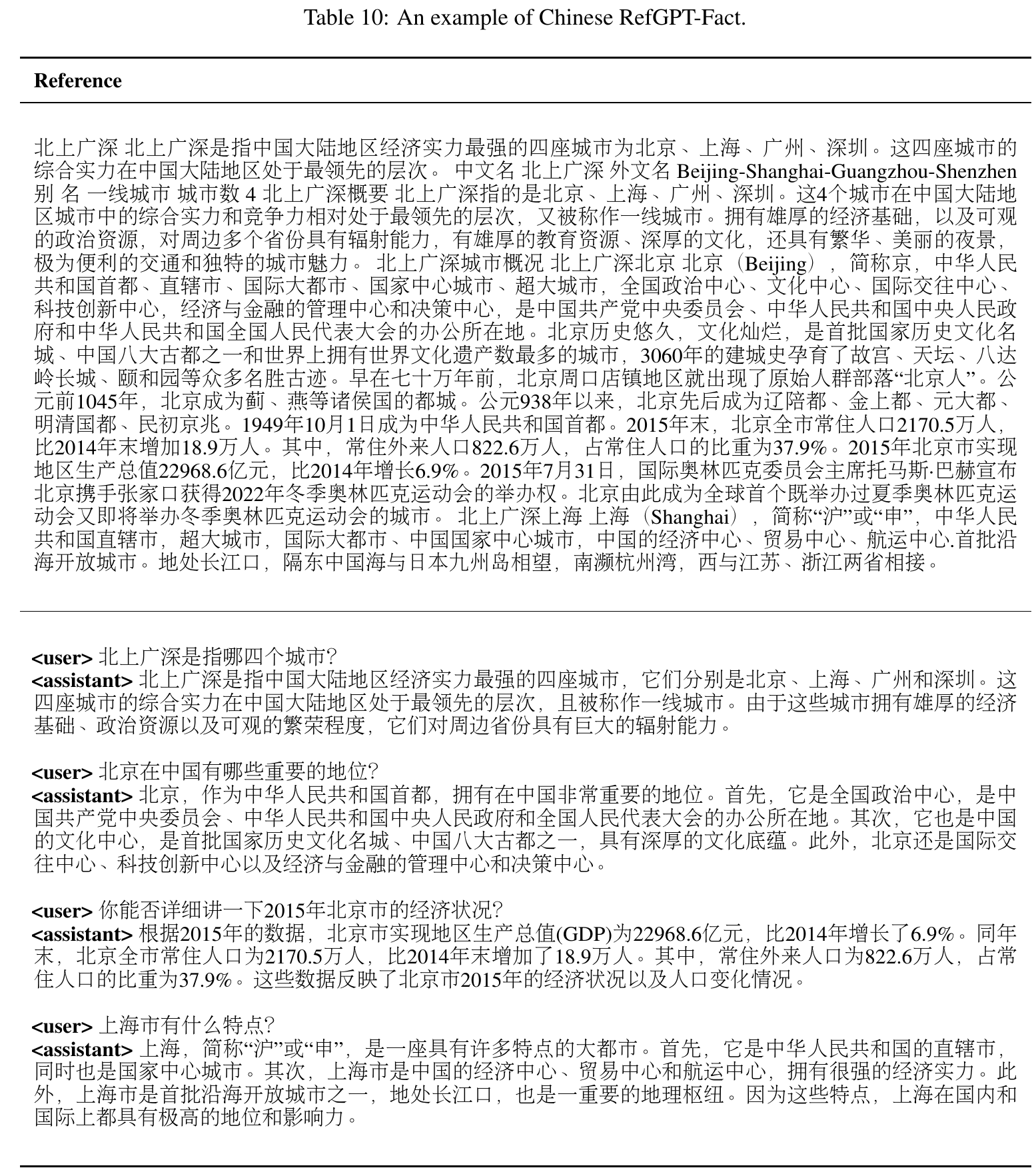}
\end{figure*}

\begin{figure*}[h]
	\centering
	\includegraphics[width=\textwidth]{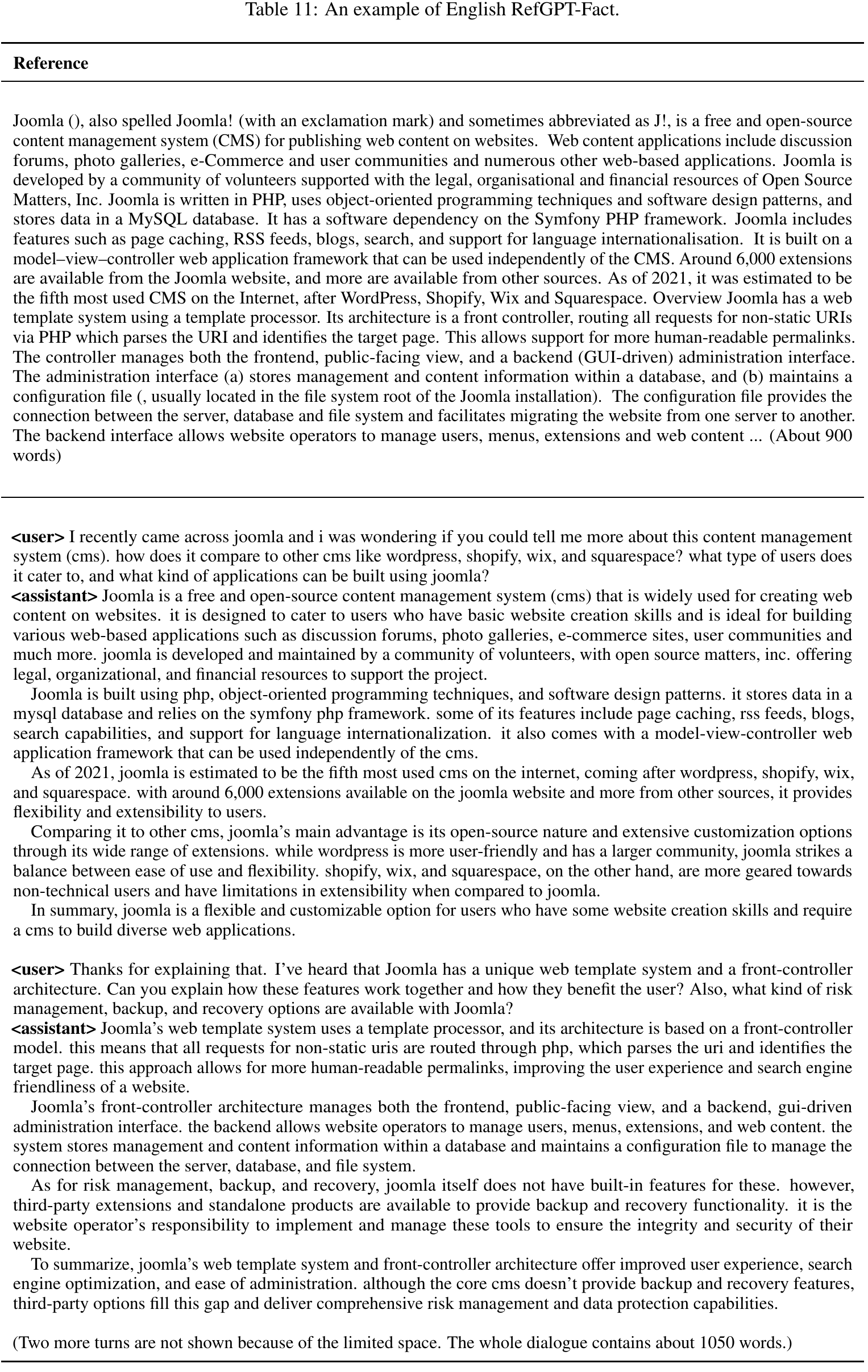}
\end{figure*}

\begin{figure*}[h]
	\centering
	\includegraphics[width=\textwidth]{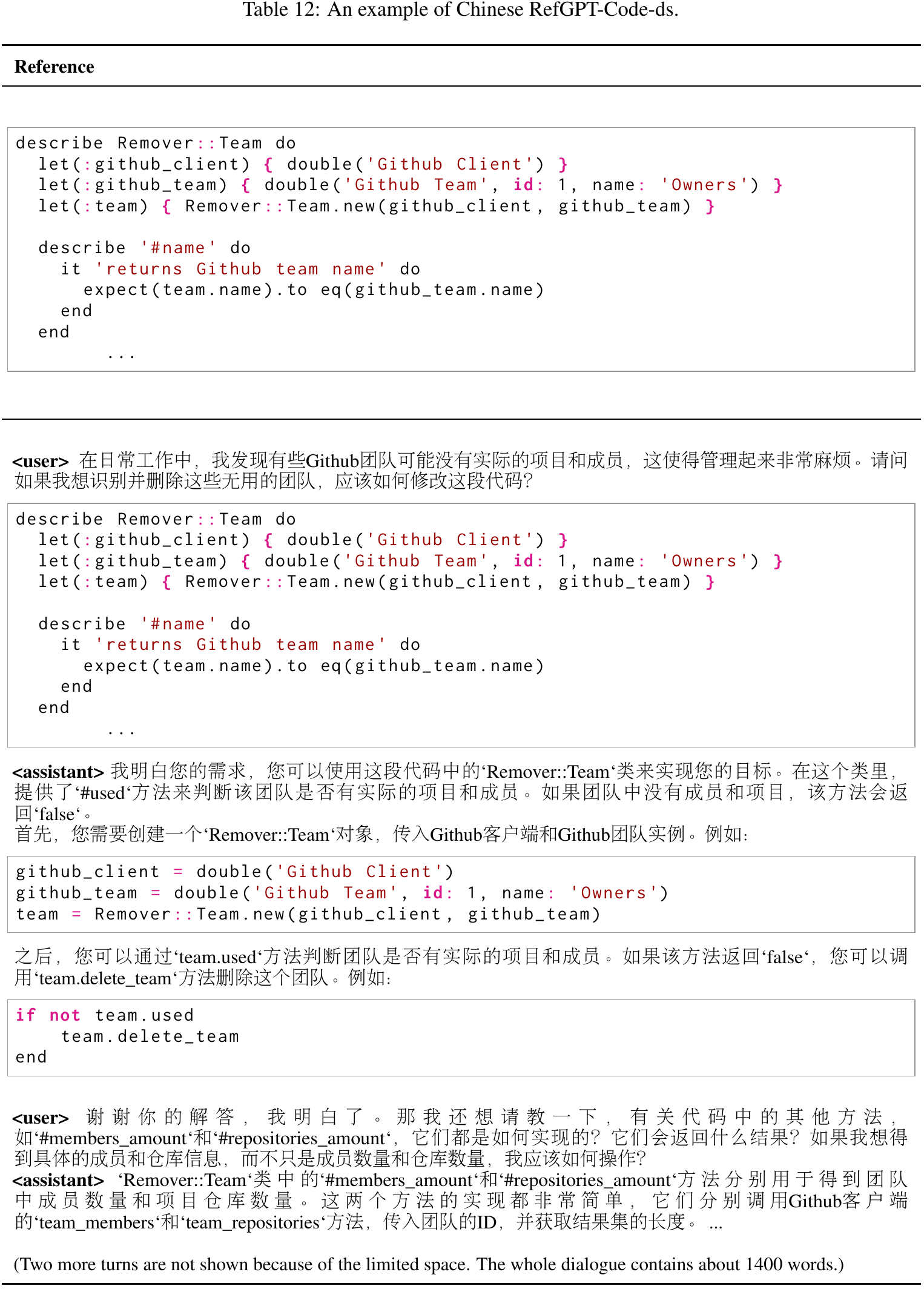}
\end{figure*}

\begin{figure*}[h]
	\centering
	\includegraphics[width=\textwidth]{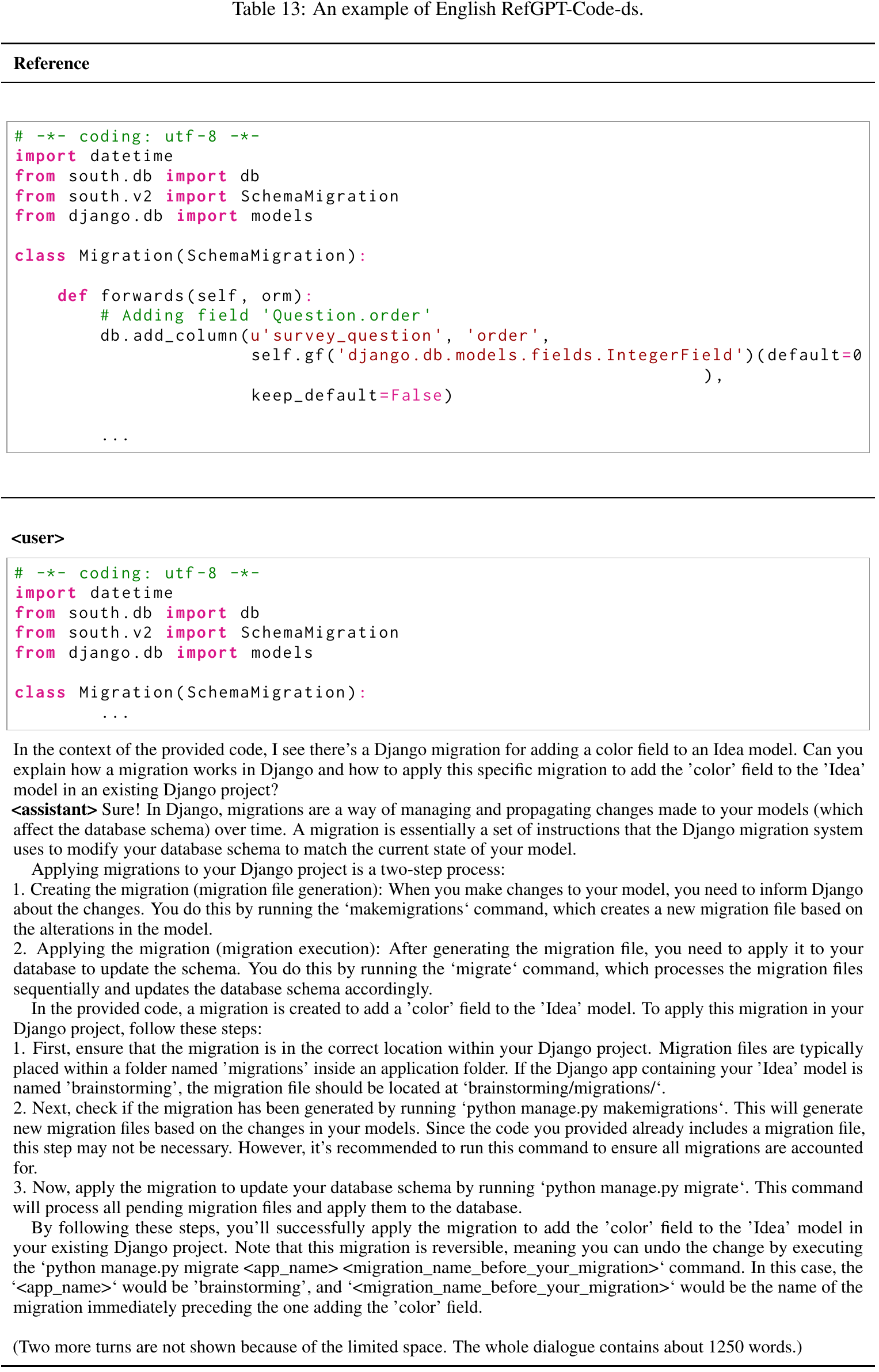}
\end{figure*}

\begin{figure*}[h]
	\centering
	\includegraphics[width=\textwidth]{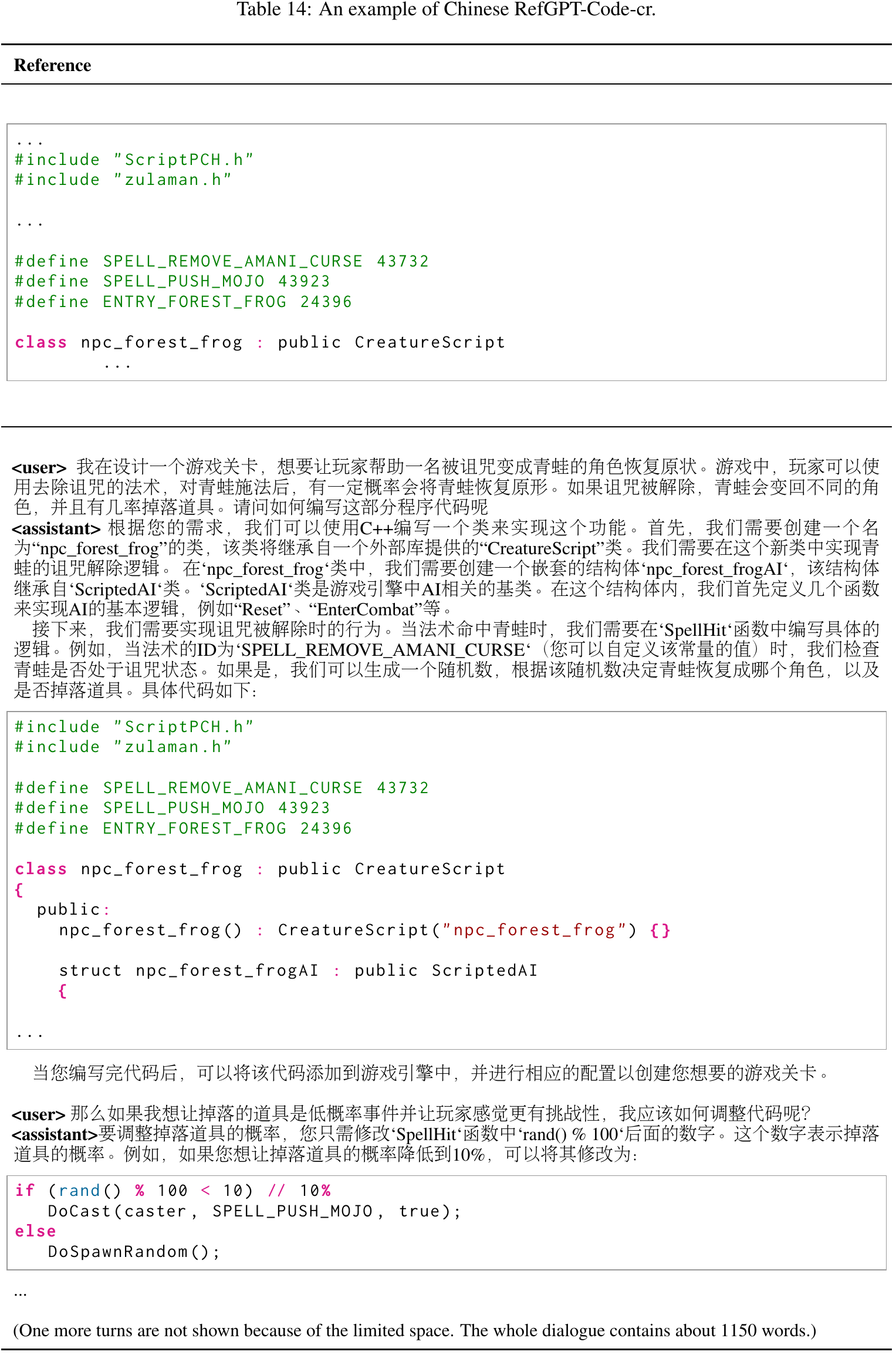}
\end{figure*}

\begin{figure*}[h]
	\centering
	\includegraphics[width=\textwidth]{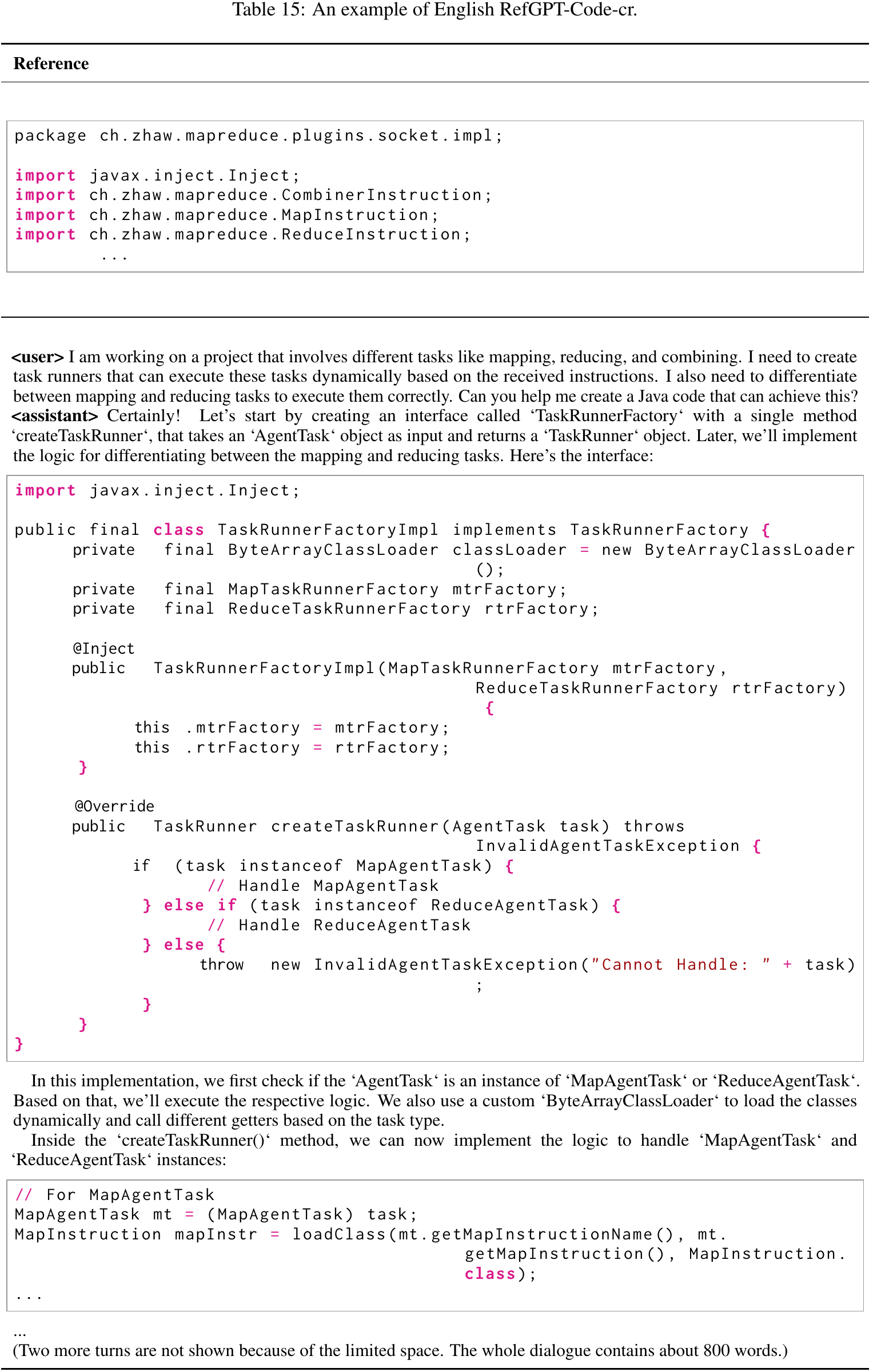}
\end{figure*}

\begin{figure*}[h]
	\centering
	\includegraphics[width=\textwidth]{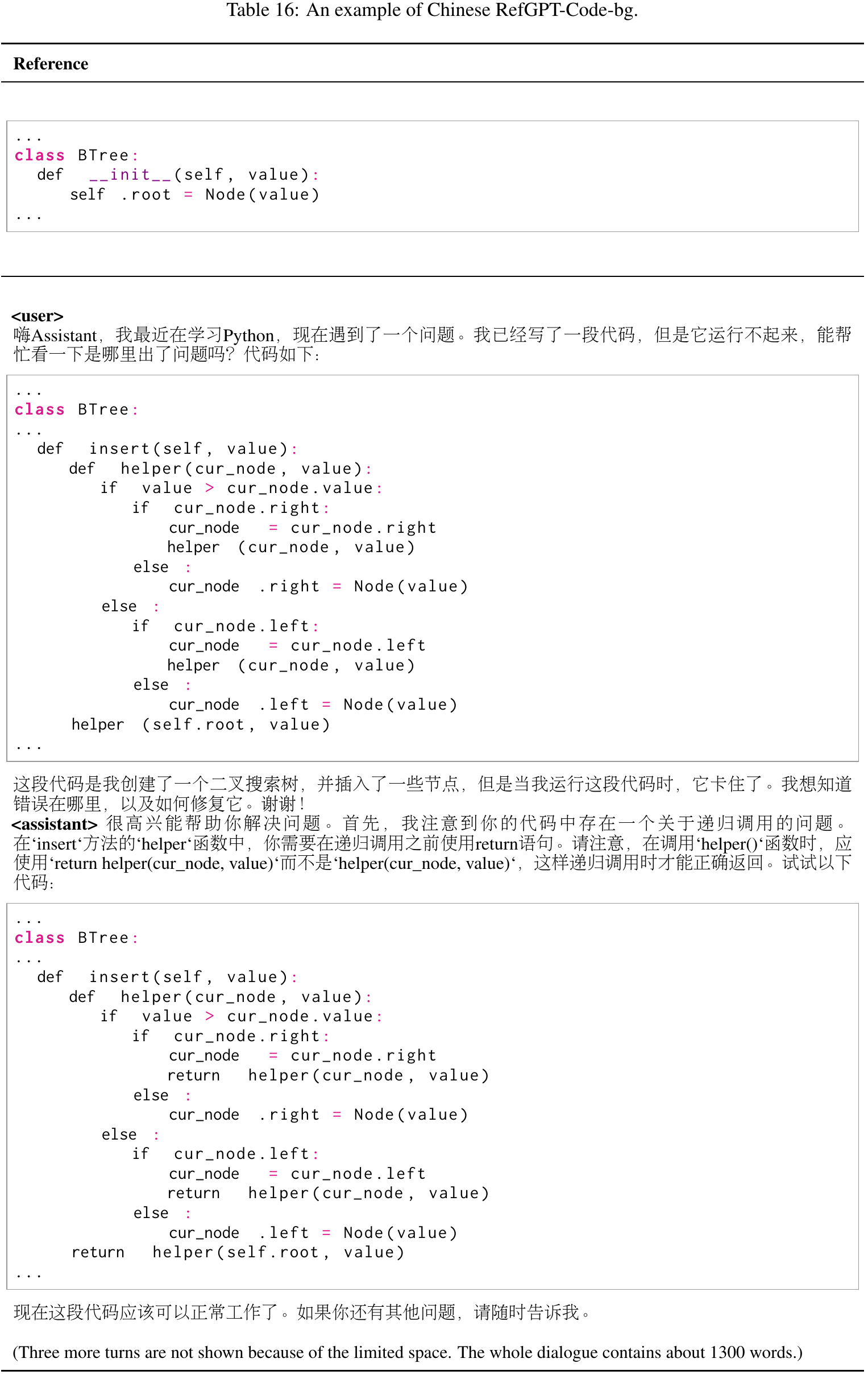}
\end{figure*}

\begin{figure*}[h]
	\centering
	\includegraphics[width=\textwidth]{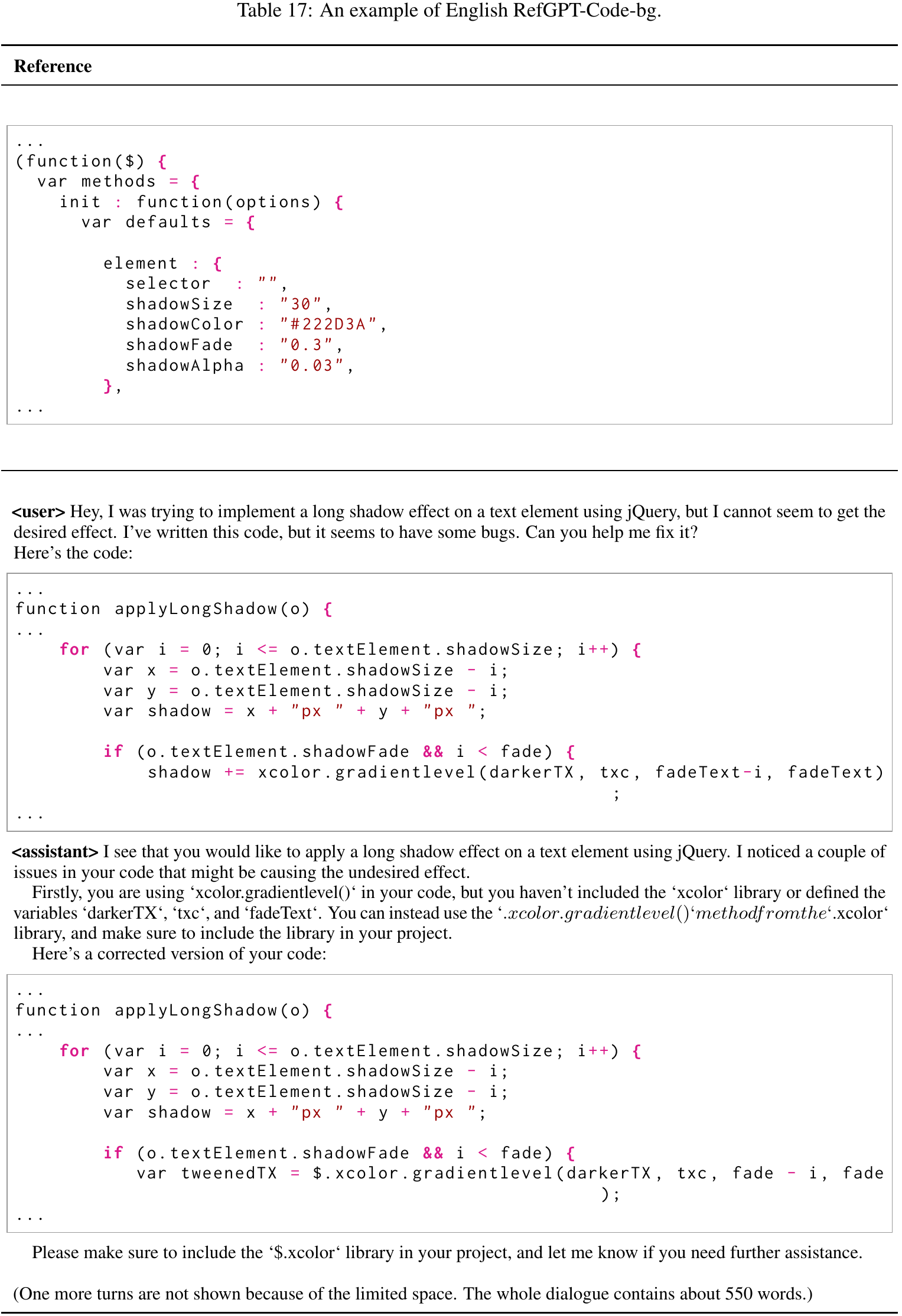}
\end{figure*}

\clearpage
\section{Truthfulness Evaluation Examples}
\label{app:truthfulness}

\begin{figure*}[h]
	\centering
	\includegraphics[width=\textwidth]{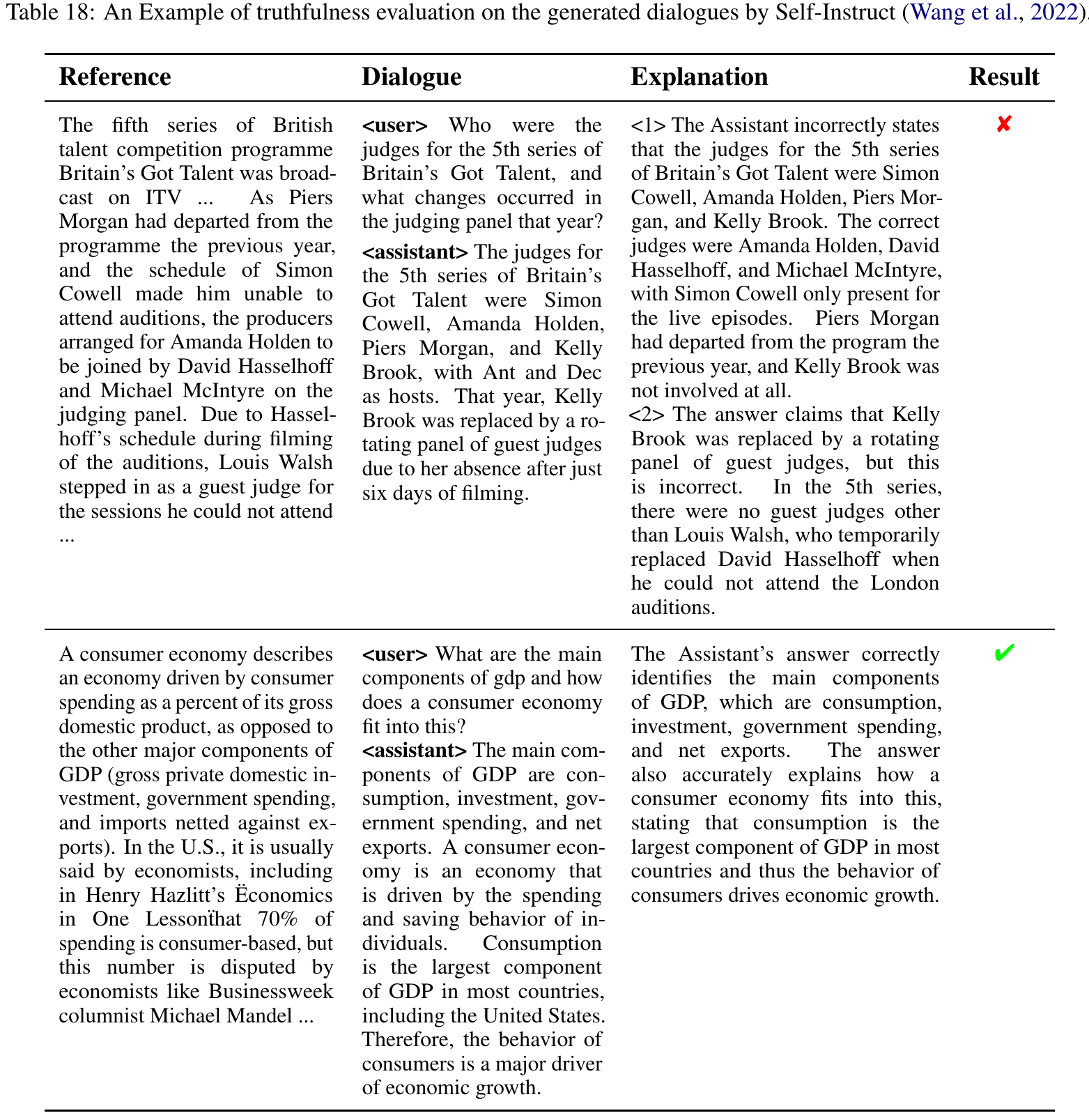}
\end{figure*}

\begin{figure*}[h]
	\centering
	\includegraphics[width=\textwidth]{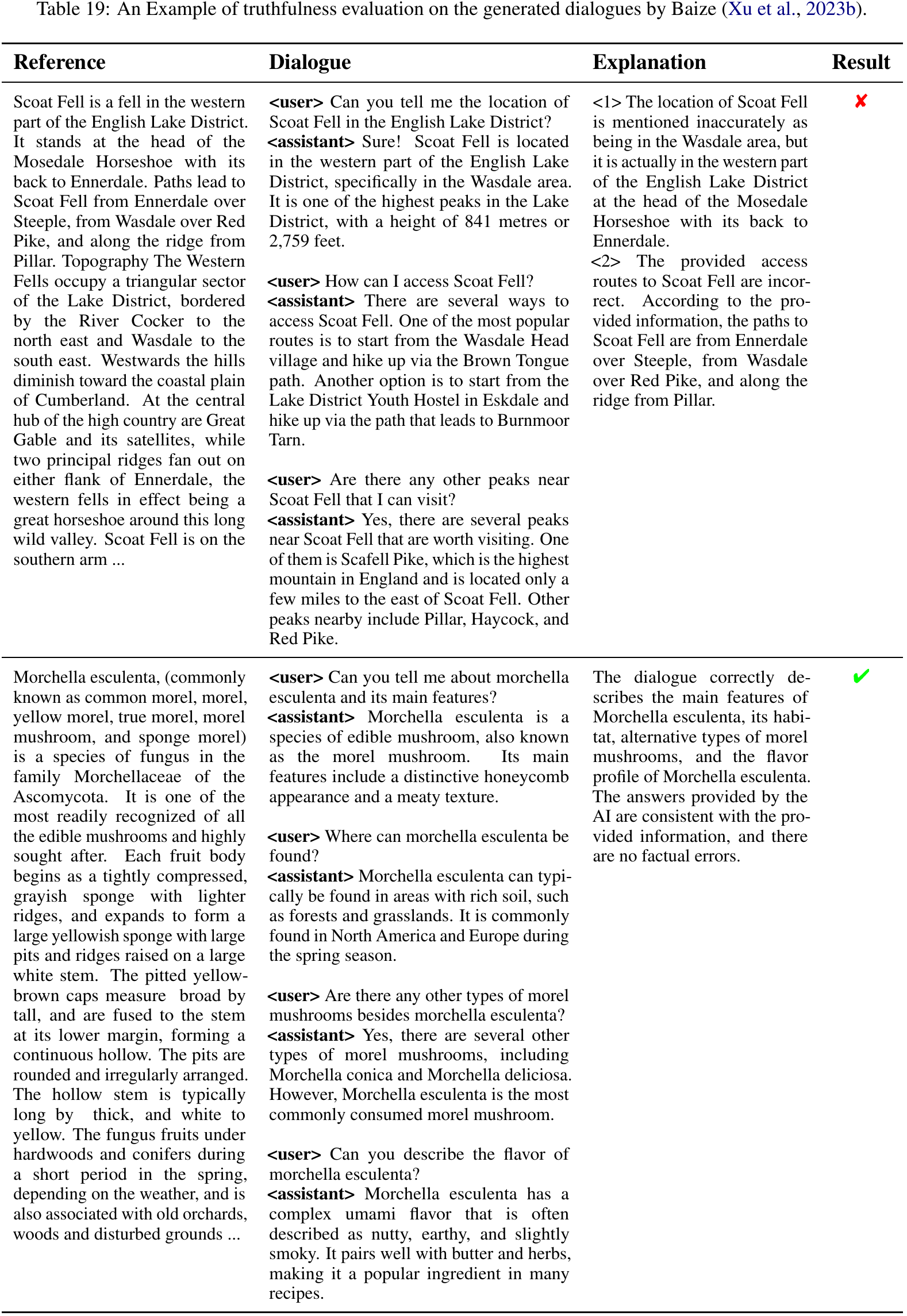}
\end{figure*}
\end{document}